\documentclass[lettersize,journal]{IEEEtran}

\usepackage{amsmath}
\usepackage{tikz}
\usepackage{subcaption}
\usepackage {color, graphicx}
\usepackage{amssymb}
\usepackage[normalem]{ulem} 
\usepackage{rotating}
\usepackage{ifthen}
\usepackage{amssymb}
\usepackage{nicefrac}
\usepackage{pifont}
\usepackage{color, colortbl}	
\usepackage{lineno}
\usepackage{booktabs}
\usepackage{multirow}
\usepackage{array}
\usepackage{dsfont}
\usetikzlibrary{3d}
\usetikzlibrary{shapes.symbols, arrows.meta}
\usetikzlibrary{decorations, decorations.pathreplacing, decorations.pathmorphing, arrows.meta, shapes.geometric, decorations.markings}
\usetikzlibrary{chains,scopes,positioning,backgrounds,shapes,fit,shadows,calc}
\usetikzlibrary{matrix,arrows, shapes.geometric, patterns}
\usetikzlibrary{patterns.meta}
\usepackage{siunitx}

\usepackage{pgfplots}
\pgfplotsset{compat=1.18}
\usepgfplotslibrary{groupplots}
\usepackage{pgfplotstable}
\usepgfplotslibrary{fillbetween}
\usepackage{siunitx}

\tikzstyle{input} = [draw, rectangle, minimum height=0.8cm, minimum width=2cm]
\tikzstyle{embedding} = [draw, rectangle, rounded corners, minimum height=0.8cm, minimum width=2cm, fill=blue!10]
\tikzstyle{addnorm} = [draw, rectangle, minimum height=1cm, minimum width=2cm, fill=green!10]
\tikzstyle{attention} = [draw, rectangle, minimum height=1cm, minimum width=2cm, fill=yellow!20]
\tikzstyle{feedforward} = [draw, rectangle, minimum height=1cm, minimum width=2cm, fill=red!10]
\tikzstyle{output} = [draw, rectangle, minimum height=0.8cm, minimum width=2cm, fill=blue!10]
\tikzstyle{arrow} = [thick,->,>=stealth]

\pgfdeclareplotmark{mydiamond}{
    \node[diamond,minimum size=6pt,
          inner sep=0pt,draw=black,solid,fill=red] {};
}

\pgfdeclareplotmark{mytriangle}{
    \node[circle,minimum size=3pt,
          inner sep=0pt,draw=black,solid,fill=green] {};
}

\pgfdeclareplotmark{mysquare}{
    \node[minimum size=4pt,
          inner sep=0pt,draw=black,solid,fill=blue] {};
}

\usepackage{setspace}

\tikzdeclarepattern{
  name=mylines,
  parameters={
      \pgfkeysvalueof{/pgf/pattern keys/size},
      \pgfkeysvalueof{/pgf/pattern keys/angle},
      \pgfkeysvalueof{/pgf/pattern keys/line width},
  },
  bounding box={
    (0,-0.5*\pgfkeysvalueof{/pgf/pattern keys/line width}) and
    (\pgfkeysvalueof{/pgf/pattern keys/size},
0.5*\pgfkeysvalueof{/pgf/pattern keys/line width})},
  tile size={(\pgfkeysvalueof{/pgf/pattern keys/size},
\pgfkeysvalueof{/pgf/pattern keys/size})},
  tile transformation={rotate=\pgfkeysvalueof{/pgf/pattern keys/angle}},
  defaults={
    size/.initial=5pt,
    angle/.initial=45,
    line width/.initial=.4pt,
  },
  code={
      \draw [line width=\pgfkeysvalueof{/pgf/pattern keys/line width}]
        (0,0) -- (\pgfkeysvalueof{/pgf/pattern keys/size},0);
  },
}

\tikzset{pics/fake box/.style args={
#1 with dimensions #2 and #3 and #4 and #5 and #6 and #7}{
    code={
       \draw[gray,ultra thin,fill=#1] (0,0,0) coordinate(-front-bottom-left) to ++ (0,#3,0) coordinate(-front-top-right) --++ (#2,0,0) coordinate(-front-top-right) --++ (0,-#3,0) coordinate(-front-bottom-right) node[right,color=black,midway,inner sep=0pt] {\scriptsize #7} -- cycle;
       \draw[gray,ultra thin,fill=#1] (0,#3,0) --++  (0,0,#4) coordinate(-back-top-left) --++ (#2,0,0)  coordinate(-back-top-right) --++ (0,0,-#4)  -- cycle; 
       \draw[gray,ultra thin,fill=#1!80!black] (#2,0,0) --++ (0,0,#4) --++ (0,#3,0) --++ (0,0,-#4) coordinate(-back-bottom-right) node[right,color=black,midway,inner sep=0pt] {\scriptsize #6} -- cycle;
       \draw[gray,ultra thin,fill=#1] (0,0,#4) --++ (0,#3,0) --++ (#2,0,0) node[above,color=black,midway,inner sep=1pt] {\scriptsize #5} --++ (0,-#3,0) -- cycle; 
    }
  }
}

\tikzset{
        line/.style={
            draw, ->, rounded corners = 3mm,
        }
}

\tikzset{
    *|/.style={
        to path={
            (perpendicular cs: horizontal line through={(\tikztostart)},
                                 vertical line through={(\tikztotarget)})
            -- (\tikztotarget) \tikztonodes
        }
    }
}

\definecolor{matblue}{RGB}{92, 107, 192}
\definecolor{matred}{RGB}{239, 83, 80}
\definecolor{matgreen}{RGB}{67, 160, 71}
\definecolor{matdarkblue}{RGB}{26, 35, 126}
\definecolor{matdarkred}{RGB}{198, 40, 40}
\definecolor{matlightblue}{RGB}{159, 168, 218}
\definecolor{matlightred}{RGB}{239, 154, 154}

\usepackage[ruled,linesnumbered,vlined]{algorithm2e}

\usepackage{cite}

\usepackage{graphicx}
\usepackage{float}

\usepackage{listings}
\usepackage{xcolor}

\usepackage{acronym}

    \acrodef{AUC}{area under the curve}
    \acrodef{COTS}{commercial off-the-shelf}
    \acrodef{EER}{equal error rate}
    \acrodef{CNN}{convolutional neural networks}
    \acrodef{FMR}{false match rate}
    \acrodef{FNMR}{false non-match rate}
    \acrodef{MSE}{mean squared error}
    \acrodef{NAS}{neural architecture search}
    \acrodef{NIST}{National Institute of Standards and Technology}
    \acrodef{SDK}{source development kit}
    \acrodef{FVC2002}{Fingerprint Verification Competition 2002 DB1 A}
    \acrodef{FVC2004}{Fingerprint Verification Competition 2004 DB1 A}

	\usepackage{url}

\begin{document}
\title{\mbox{Infant Fingerprints: Quality Assessment \& Recognition}} 

\author{João L. H. Dall Agnol, Dalcimar Casanova, Rodrigo Minetto, \\Luiz F. P. Southier, Marcelo Teixeira, Marcelo Filipak, and Anil K. Jain

\thanks{João Leonardo Harres Dall Agnol, Dalcimar Casanova, Rodrigo Minetto, Luiz Fernando Puttow Southier, Marcelo Teixeira, and Marcelo Filipak are with the Federal University of Technology - Paraná (UTFPR), Brazil (e-mails: joaoleonardoagnol@alunos.utfpr.edu.br, dalcimar@utfpr.edu.br, rminetto@utfpr.edu.br, luizsouthier@utfpr.edu.br, mtex@utfpr.edu.br, and mfilipak@alunos.utfpr.edu.br).}
\thanks{Anil Kumar Jain is with Michigan State University, East Lansing, MI, USA (e-mail: jain@cse.msu.edu).}}

	\maketitle

	\begin{abstract}
        Infant fingerprint recognition is hindered by a combination of physiological characteristics, such as small ridge width and spacing, and acquisition challenges that include involuntary motion, skin residue, and uneven fingertip pressure. These factors undermine fingerprint quality standards originally designed for adults, which assume feature distributions that do not hold for infants. To address this gap, this paper evaluates ten fingerprint quality assessment methods on a longitudinal dataset that spans six age groups from newborns to three months, alongside a parameter tuning routine that uncovers age-specific fingerprint characteristics to recalibrate methods with domain-dependent assumptions. Both default and tuned configurations are compared through recognition experiments assessing the effectiveness of quality-guided gallery selection. Template-fusion strategies and latent fingerprint recognition technologies are further examined to identify the best possible matching rates achievable by the top-performing method. Results show that tuned parameter sets outperform default configurations for the majority of methods and age groups, indicating that targeted recalibration of existing frameworks improves gallery utility and contributes to more reliable infant fingerprint recognition.
	\end{abstract}


\section{Introduction}~\label{sec:introduction}

\begin{figure}[!htb]
    \centering
    \input{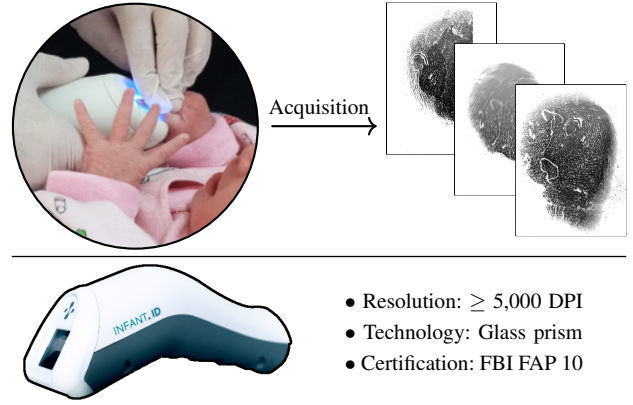}
    \caption{Infant fingerprint acquisition process: top-left shows capture procedure used for the UTFPR-IFD dataset~\cite{Southier2025:base}; top-right contains examples of fingerprint video frames obtained from a single newborn fingerprint; bottom shows the InfantID scanner used to acquire the images, alongside its key characteristics such as sampling rate and technology~\cite{InfantID2024}.}
    \label{fig:intro-utfprifd}
\end{figure}

Infant and adult fingerprints exhibit substantially different structural patterns. Infant ridges are markedly thinner and inter-ridge spacing may be up to four times smaller~\cite{Jain2017, Saggese2019}, which directly motivates the use of high-resolution (1200-5000 PPI) scanners to preserve sufficient discriminative detail for recognition \cite{Jain2017, Saggese2019, Moolla2021, Engelsma2022, Southier2025:base}. In addition, infant fingers are softer and more elastic than those of adults, increasing susceptibility to deformation during image capture \cite{Moolla2021}. Figure~\ref{fig:intro-utfprifd} illustrates the acquisition process, alongside sample images that exhibit common capture degradations, including motion and residual contaminants such as lint and fiber residues.

Solutions aiming to provide infants with reliable recognition systems must grapple with these physiological challenges; particular to this work is the problem of defining an acceptance criteria in structural conformity to quantify the quality of an infant fingerprint sample. In other words, the properties that constitute an acceptable sample for adult fingerprints may not hold for infants, and the structural bounds that define ridge and valley conformity must be reconsidered according to the subject's age.

Quality assessment plays a critical role within the biometric pipeline. By bridging the capture and processing stages, it influences the reliability of all subsequent operations. For adult biometrics, fingerprint quality assessment has been extensively investigated over the past decades, leading to the development of standardized methodologies such as ISO/IEC 29794-1~\cite{ISO} and the widely adopted National Institute of Standards and Technology (NIST) Fingerprint Image Quality 2 (NFIQ2) package \cite{Tabassi2022}, which is integrated into modern comparators such as VeriFinger \cite{Neurotechnology2025}. In contrast, quality assessment for infants is a relatively unexplored area, remaining largely restricted to the direct application of NFIQ (1 and/or 2) to infant fingerprint data \cite{Jain2017, Camacho2017, Preciozzi2020}. Recent works on the topic of infant biometrics have focused on producing alternative high-resolution scanners \cite{Saggese2019, Moolla2021} and the development of image enhancement techniques, including binarization \cite{Shi2021, Harres2025} and super-resolution methods \cite{Engelsma2022, Machado2025}.

To address fingerprint quality assessment for infants, this study reviews established ISO/IEC 29794-1~\cite{ISO} methods and more recent advancements \cite{Tabassi2022, Hendre2022}, selected according to their primary underlying features (e.g., ridge spacing, orientation), to determine whether existing solutions can reliably identify informative infant fingerprint samples, and explores age-sensitive parameter search to improve recognition rates. Both standard and tuned configurations are evaluated through a gallery selection problem: given a set of images from an identity, methods must rank samples according to their matching potential --- a property commonly formalized in terms of utility for identification systems~\cite{Grother2007}. Experiments are repeated for six age-based subsets, spanning from newborns to three months of age, drawn from the UTFPR-IFD~\cite{Southier2025:base} longitudinal infant database.

The main contributions of this paper are as follows:

\begin{itemize}
\item[(i)] We review and evaluate ten fingerprint quality assessment methods, encompassing classical signal-processing, structure-driven, and learning-based approaches, that capture complementary fingerprint properties through both local and global image features.
\item[(ii)] A systematic parameter search routine is introduced to identify optimal infant ridge and valley property (width, distance, etc) ranges. This process is performed independently for each parameter-dependent method and age group, aiming to maximize the biometric utility of galleries selected through quality ranking.
\item[(iii)] To assess the effectiveness of quality-driven fingerprint selection across age groups, identity galleries are composed using the top-ranked samples selected by each quality assessment method, separately. Using VeriFinger 2025.2~\cite{Neurotechnology2025}, quantitative and qualitative analyses 
compare standard and optimized parameter configurations for infant fingerprints.
\item[(iv)] A complementary analysis examines a practical deployment scenario, combining a template-fusion strategy with a state-of-the-art latent comparator (MegaMatcher 2025.2~\cite{Neurotechnology2025}) to investigate the best possible matching rates achieved by a top-performing quality method identified in this study.
\end{itemize}

To the best of the authors’ knowledge, this is the first study to evaluate and tune fingerprint quality assessment methods in the context of infant biometrics. 

This paper is structured as follows: Section~\ref{sec:lit} reviews related work, including quality assessment methods evaluated for parameter optimization in infant fingerprint recognition, as described in Section~\ref{sec:met}. Section~\ref{sec:res} evaluates the impact of these methods on recognition performance, with and without tuned parameters; finally, Section~\ref{sec:dis} discusses the implications of these findings, and Section~\ref{sec:con} summarizes the contributions and outlines future research directions.

\section{Fingerprint quality estimation methods}\label{sec:lit}

Fingerprint quality estimation has been addressed through a broad range of methodologies that differ both in features used to characterize ridge patterns (orientation, width, frequency, pores, minutiae) and in how such features are extracted, from handcrafted descriptors to deep learning representations. Existing surveys~\cite{Fernandez2007, Olsen2016, Jain2022} focus on adult fingerprints, leaving infant fingerprint quality assessment largely unexplored.

\begin{figure*}[!htb]
    \centering
    \input{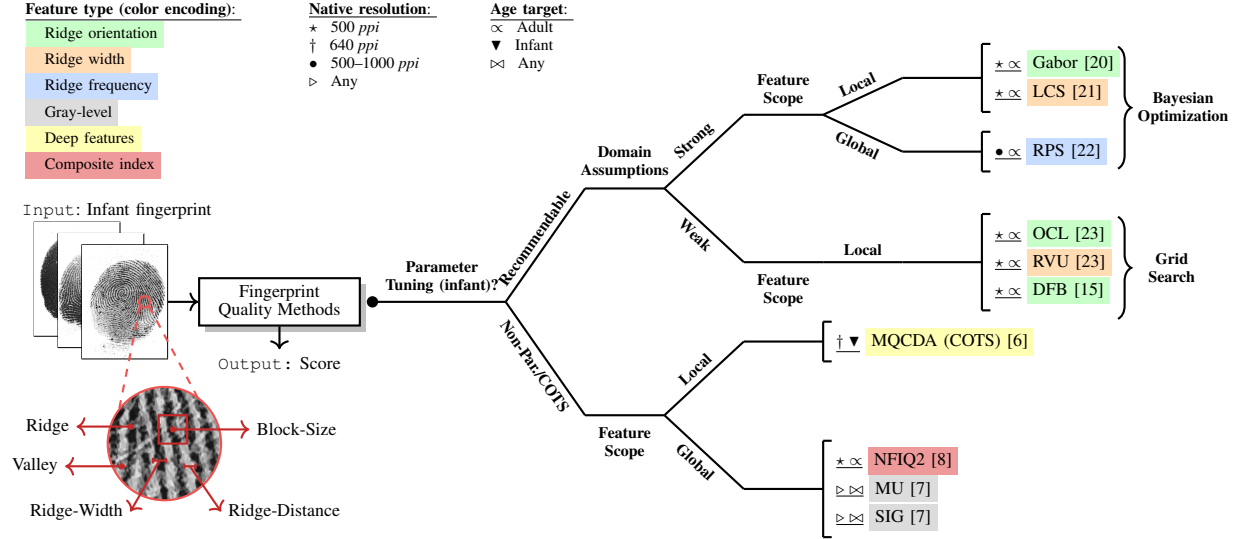}
    \caption{Fingerprint quality estimation taxonomy. Methods are classified according to their primary feature, resolution and the age they target. Features directly related to visible patterns are also showcased.}
    \label{fig:lit-schema}
\end{figure*}

Figure~\ref{fig:lit-schema} organizes the methods reviewed here into a taxonomy structured around their adaptability to infant fingerprints, degree of parameterization, dependence on domain-specific assumptions, and feature scope. From left to right, the first division in the taxonomy distinguishes
between methods that can be explicitly tuned to infant fingerprints through parameter optimization, and those that rely on fixed configurations, learned models, or intrinsic image statistics. This distinction is particularly relevant because infants exhibit substantially different ridge spatial characteristics compared to adults \cite{Camacho2017, Preciozzi2020, Moolla2021, Engelsma2022}. 

Methods subjected to parameter optimization are grouped into two categories:
\begin{itemize}
    \item \emph{Strong domain assumptions}: methods whose parameters are directly associated to fingerprint morphology, including ridge frequency, ridge width, and orientation consistency. Representative examples include Gabor-based methods~\cite{Shen2001}, gray-level analysis for Local Clarity Score (LCS~\cite{Chen2004}), and Radial Power Analysis (RPS~\cite{Chen2005}). Their dependence on expected ridge-valley patterns makes them particularly relevant for domain-specific optimization in infant fingerprints.
    \item \emph{Weak domain assumptions}: comprises methods --- Orientation Certanty Level (OCL~\cite{Lim2002}), Ridge-Valley Uniformity (RVU~\cite{Lim2002}) and Directional Filter Bands (DFB~\cite{Hendre2022}) --- whose parameters are exclusively related to the size of a local analysis region. Although these methods rely less on explicit assumptions about ridge morphology, their performance may still depend on the appropriate spatial scale for the finer ridge patterns found in infant fingerprints.
\end{itemize}

Methods with minimal or implicit parameterization form the remaining group: (i) purely statistical approaches, mean gray-level intensity (MU~\cite{ISO}) and the standard deviation of gray-level intensities (SIG~\cite{ISO}), that need no explicit parameters; and (ii) data-driven COTS solutions (MQCDA~\cite{InfantID2024} model, NFIQ2~\cite{Tabassi2022}) trained on fixed configurations. The simpler statistical methods, which assess quality from global grey-level intensity, are more readily applicable to infants, but are correspondingly less sensitive to fine structural detail. The current obstacle is thus to determine the expected ranges for ridge and valley thickness or dominant frequency that best suit infant fingerprints.

Sampling rate intensifies this barrier: most solutions assume a sampling rate near 500 PPI, while infant fingerprints require higher rates to resolve sufficient detail. One exception is \cite{Teixeira2017}, which uses pores as quality indicators for high-resolution images; however, infant pores are rarely visible, particularly in the first months of life. As later detailed in Section~\ref{subsec:met-data}, this study is limited to analyses of 5000 PPI images downsampled to 500 PPI to maintain compatibility with COTS comparators and spatial correspondence with adult fingerprints.

Lastly, all methods are further categorized as either global, assigning a single score to the entire image, or local, scoring regions separately. Though localized representations are generally preferred \cite{Jain2022}, global approaches remain widely adopted largely due to NFIQ2. Local scoring is, however, useful for applications such as image enhancement \cite{Turroni2012, Panetta2019} and composition \cite{Yu2025}, potentially benefiting the broader infant fingerprint recognition pipeline.

To the best of the authors’ knowledge, no existing study targets infant fingerprint quality specifically. This gap is reasonably explained by the scarcity of infant fingerprint datasets \cite{Southier2025} and by the lack of robust synthetic fingerprint generation even for adults \cite{Peng2026}. Consequently, how well current quality assessment methods transfer to infants depends heavily on their underlying assumptions and adaptability to age-related biometric differences.

The following section proposes a parameter search for reviewed methods, using a longitudinal database \cite{Southier2025:base} to capture the patterns of different infant age groups.

\section{Parameter search for infants}\label{sec:met}

Given that several quality assessment methods rely on intrinsic fingerprint features such as ridge width and spacing, a natural question arises as to whether, and to what extent, parameters originally derived from adult distributions appropriately capture infant features. To investigate this scenario, a parameter search is performed over infant subjects from hours after birth to 3 months of age.

\subsection{Infant Dataset and Experimental Protocol}\label{subsec:met-data}

\begin{figure}[!htb]
    \centering
    \input{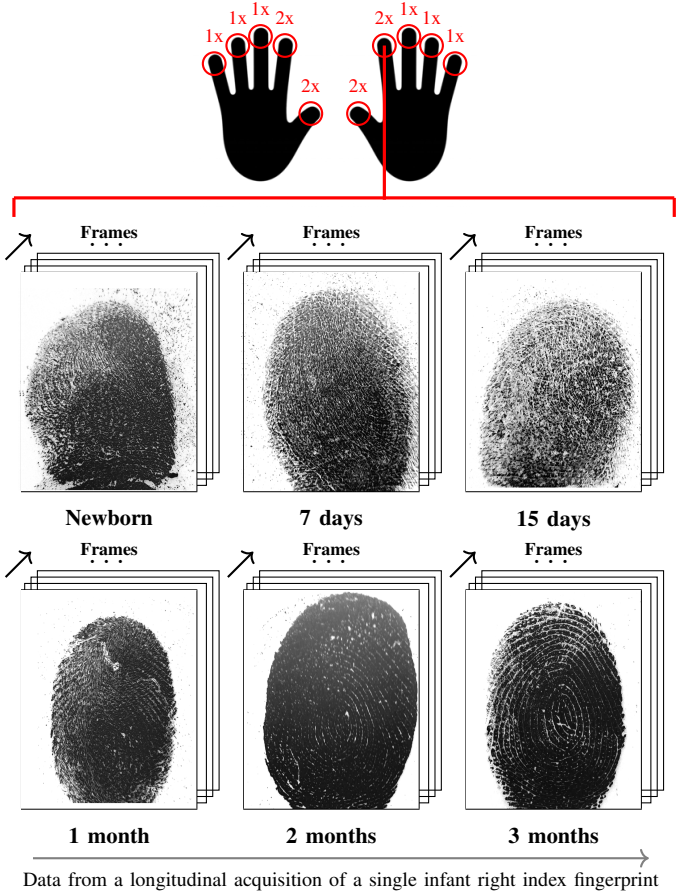}
    \caption{Overview of the UTFPR-IFD dataset~\cite{Southier2025:base}. A total of 14 fingerprint videos are acquired per session from both hands, with thumbs and index fingers captured twice due to their higher recognition utility. Stacked images represent the video-based frames collected at each stage of life
    }
    \label{fig:intro-ageing}
\end{figure}

The UTFPR infant fingerprint dataset (UTFPR-IFD)~\cite{Southier2025:base} spans longitudinal video acquisitions captured using the high-resolution ($\sim$5000 PPI) InfantID optical fingerprint scanner. The dataset includes recordings from infants at multiple age milestones, beginning with an initial acquisition in the São Lucas hospital (city of Pato Branco, Paraná state, Brazil), within 3–24 hours after birth, followed by scheduled follow-up in-house visitations (see Figure~\ref{fig:intro-ageing}). The data collection protocol was approved by the Brazilian research ethics committee under Certificate of Ethical Appreciation Presentation (CAAE) 73791023.7.0000.0177 via \emph{Plataforma Brasil}. This dataset is not currently available to the general public due to ethical restrictions related to the age of the subjects. Consent for collection was provided by their legal guardians.

This study considers a curated subset of the UTFPR-IFD, which was selected according to two inclusion criteria: (i) subjects must have complete longitudinal coverage across all acquisition milestones up to three months of age; and (ii) each session must contain the full set of 14 expected video samples, as acquisition artifacts may occasionally lead to missing data. Based on the longest currently available observation window (March 2024 to September 2025), a total of 45 infants met these criteria. In total, this subset comprises 3,780 high-resolution video sequences ($45 \times 14 \times 6$) acquired from 450 distinct digits across 6 sessions over time, and containing 135,382 video frames in total.

A 5-fold cross-validation is applied to the selected subjects, partitioning five different subsets for each age group that contains 36 subjects for training and 9 for testing. The same subject partition is applied to all six age groups to prevent identity leakage. Training data is further subsampled using a minimum inter-frame distance of three frames to reduce the temporal redundancy inherent to video-based acquisitions. This limits correlation and computational overhead while preserving sufficient variability for parameter search.

Neurotechnology's VeriFinger 2025.2 and MegaMatcher SDK 2025.2 \cite{Neurotechnology2025} are employed to evaluate frame selection performance. Because these technologies are designed to operate on 500 PPI images, which remains the de facto standard for most operational systems, prior to template extraction all video frames were downsampled from 5000 to 500 PPI using bilinear interpolation with pixel area relation, while preserving the original aspect ratio. This preprocessing step reflects a practical limitation of computation and current fingerprint matching technology, as support for high-resolution ($>$500 PPI) fingerprint matching remains limited in commercial-grade SDKs. Similar protocols have been adopted in previous studies on infant fingerprints, where high-resolution acquisitions were resampled to standard resolutions to ensure compatibility with existing matchers \cite{Saggese2019}. 

\subsection{Parameter Search Space}\label{subsec:met-parameters}

A parameter search is proposed in order to adapt existing fingerprint quality assessment methods to infant biometrics, tuning input values that directly influence the analysis of ridge structure, spatial scale, and frequency content. Methods that do not expose tunable parameters are excluded from this routine. Table~\ref{tab:method-params} summarizes the parameters considered during tuning, with their corresponding search spaces.

\begin{table}[!htb]
\centering
\scriptsize
\caption{Summary of methods subjected to parameter tuning and their corresponding search variables.}
\label{tab:method-params}
\begin{tabular}{ll}
\hline
\textbf{Method} & \textbf{Parameters tuned} \\
\hline
Gabor & $m$, $f$, $\sigma$, Gaussian window ($\sigma_\theta$) \\
LCS  & $W_r^{\min}, W_r^{\Delta}, W_v^{\min}, W_v^{\Delta}$, block size \\
RPS  & $r_{\min}, r_{\Delta}, \Delta r$, $\theta$ \\
OCL  & block size \\
RVU  & block size \\
DFB  & block size \\
\hline
\end{tabular}
\end{table}

\subsubsection{Gabor}

Assesses quality by measuring how strongly a bank of orientation- and frequency-tuned filters responds to the local ridge-valley pattern. If the frequency $f$ is mismatched to the true ridge periodicity, filter responses weaken across all orientations, flattening the response distribution and undermining orientation as a reliable quality indicator.

The search space for the frequency parameter $f$ spans $[2, 20]$ pixels, corresponding to $[0.05, 0.5]$ cycles per pixel. For reference, \cite{ISO} uses 10 pixels ($f=0.1$) for adults. The Gaussian deviation $\sigma$, which defines the spatial extent of the Gaussian envelope, is not tuned directly, but rather computed from the chosen $f$ using
\begin{equation}
\sigma = \frac{1.5}{\sqrt{6 \ln(10)}} \,p
\end{equation}
where $p$ is the spatial frequency in cycles per pixel. This parameterization links the envelope width to the ridge spacing, maintaining a consistent filter scale as the ridge period varies. The number of filter orientations $m \in [4, 8]$ and the Gaussian window deviation $\sigma_\theta \in [1, 10]$ are also tuned.

\subsubsection{LCS}

Evaluates fingerprints based on the clarity of ridge-valley separation, over image blocks, using the average widths of ridges ($\overline{W_r}$) and valleys ($\overline{W_v}$). As such, two width intervals are tuned: the ridge-width bounds $(W_r^{\min}, W_r^{\max})$ and the valley-width bounds $(W_v^{\min}, W_v^{\max})$. 

To reduce redundancy in the search space, the upper bound of each interval is encoded as a displacement $W_{\max} = W_{\min} + W_{\Delta}$, with bounds ranging from $1$ to $20$ pixels. This search space encompasses and extends the nominal adult ranges specified in \cite{ISO} for 500 PPI fingerprints, accommodating narrower valleys and relative increase in ridge width observed in infant fingerprints downsampled to this scale.

\subsubsection{RPS}

Applies the Discrete Fourier Transform (DFT) to quantify the concentration of energy around the dominant ridge frequency. The parameters that represent the bounds of this dominant frequency, $r_{\min}$ and $r_{\max}$, adopt the same displacement encoding used for LCS, with $r_{\max}=r_{\min}+r_{\Delta}$, relative to the expected image dimensions at 500 PPI. Accordingly, the search space is defined by $r_{\min}$ and $r_{\Delta}$, both constrained to the interval $[0.01,0.69]$. This range extends well beyond the reference value ($r_{\min}\approx 0.077$ \cite{ISO}), covering frequencies from $0.01$ (seven times lower than an adult) up to $0.70$ (the maximum admissible frequency), while ensuring that $r_{\max}$ remains greater than $r_{\min}$.

The remaining parameters, radial distance $\Delta r$ and angular resolution $\theta$, are tuned over $[1, 100]$, spanning one order of magnitude below and above the adult standard of $10$~\cite{ISO}, and $[60, 180]$ degrees, respectively.

\subsubsection{Grid-search methods}

OCL, RVU, and DFB are not dependent on external bounds or parameters besides the size of local image blocks. These values are evaluated via grid-search, alongside LCS own block size parameter, ranging from $4\times4$ to $128\times128$ pixels. For RVU and LCS, rectangular blocks (aligned to ridge orientation) are updated accordingly and require no direct optimization.

Although these values are not directly related to fingerprint structures such as ridge width and spacing, larger block sizes can reduce computation time, and may also improve robustness for infant fingerprints since broader regions capture more ridges and valleys, which can stabilize orientation estimation in OCL and DFB and yield more representative width averages in RVU, particularly given the lower local clarity of infant fingerprint images.

Block size of $256\times256$ and above were excluded, as they would cover approximately one quarter of the total fingerprint area at 500 PPI, compromising local analysis. DFB requires an additional evaluation of all multiples of $2$ from $4$ to $128$, as its base block size is $4\times4$.

\subsection{Parameter Optimization}\label{subsec:met-opt}

The parameter optimization procedure depends on the dimensionality of each method's parameter space (Figure~\ref{fig:lit-schema}). Grid search was used only for block size, a single-parameter space small enough to evaluate exhaustively. Bayesian Optimization~\cite{Nogueira2014} was employed for methods with multiple parameters, where exhaustive search is impractical, because it uses previous observations to choose next evaluations.

False Non-Match Rate (FNMR) was chosen as the loss for both strategies because sample quality is more strongly associated with genuine comparisons than with impostor ones \cite{Grother2007}. A high-quality genuine pair is expected to match, whereas an impostor match can occur at any quality level through incidental similarity between different fingerprints. FNMR (Equation~\ref{eq:fnmr}) is the fraction of genuine pairs $(i,j)$ in the set $\mathcal{G}_x$ whose comparison score $s(i,j)$ falls below the decision threshold $\tau$:

\begin{equation}
\mathrm{FNMR}(x, \tau) = \frac{1}{|\mathcal{G}_x|} \sum_{(i,j)\in\mathcal{G}_x} \mathds{1}\left[ s(i,j) < \tau \right]
\label{eq:fnmr}
\end{equation}

Quality assessment defines $\mathcal{G}_x$ by discarding the $20\%$ of genuine pairs with the lowest combined quality, where the combined quality of a pair is the minimum of the quality scores of its two images under configuration $x$. A $20\%$ discard rate is typical in operational biometric systems \cite{Schlett2024}. The threshold $\tau$ is fixed as the value at which $80\%$ of the genuine pairs in the initial training set $\mathcal{G}$ are matched (i.e., $\mathrm{FNMR}=20\%$ before discarding). Under this threshold, a quality method that better predicts biometric utility discards more of the false non-matches, yielding a lower FNMR on $\mathcal{G}_x$.

As such, Bayesian optimization maximizes the complement of FNMR (Equation~\ref{eq:fx-fnmr}), which is equivalent to minimizing FNMR (Equation~\ref{eq:bayes-opt}):

\begin{equation}
f(x) = 1 - \mathrm{FNMR}(x, \tau)
\label{eq:fx-fnmr}
\end{equation}

\begin{equation}
x^* = \operatorname*{arg\,max}_{x \in \mathcal{X}} f(x)
\label{eq:bayes-opt}
\end{equation}

This work utilized $10$ initial observations and $20$ subsequent iterations ($30$ evaluations in total). Up to $100$ iterations were tested, but $30$ evaluations gave comparable results while limiting overfitting to the training set. Grid search evaluated block sizes from $4$ to $128$ in powers of $2$.

\section{Experimental Results}\label{sec:res}

This section presents the parameter tuning results and extends the fingerprint comparison analysis to True and False Matching Rates (TMR@FMR), encompassing all methods across the observed age groups. A complementary analysis examines template fusion with a latent comparator to improve matching rates.

\subsection{Parameter Tuning}

Table~\ref{tab:bayes-parameters} summarizes the final parameter intervals for relevant fingerprint characteristics, namely the dominant ridge frequency (RPS) and the ridge/valley width parameters used by LCS. The reported intervals were obtained over in a separate execution that used the full 450 digits dataset to obtain a representative characterization of the entire available cohort. Although these quantities are used as algorithmic parameters in fingerprint enhancement and quality assessment, they are closely related to underlying ridge-valley features and therefore provide insights into the image properties that mostly influence matching performance across different ages.

\begin{table}[!htb]
\caption{Tuned RPS and LCS parameter intervals for each age group (in pixels). Adult reference ranges reported in \cite{ISO} are included for comparison.}
\centering
\scriptsize
\begin{tabular}{l c c c}
\hline
\textbf{Age} & \textbf{Dom. Frequency} & \textbf{Rid. Width} & \textbf{Val. Width} \\
\hline
Newborn & 0.211 - 0.397 & 2-11 & 1-3 \\
7-day & 0.379 - 0.560 & 3-6 & 1-3 \\
15-day & 0.440 - 0.700 & 2-14& 2-3 \\
1-month & 0.240 - 0.700 & 2-9 & 2-5 \\
2-month & 0.440 - 0.700 & 2-9 & 2-5 \\
3-month & 0.440 - 0.700 & 2-9 & 1-6 \\
\hline
Adult & 0.077 - 0.143~\cite{ISO} & 3-10~\cite{ISO} & 2-10~\cite{ISO} \\
\hline
\end{tabular}
\label{tab:bayes-parameters}
\end{table}

The dominant ridge frequency is algorithmically found to be consistently higher for infants. It is also shown to increase with age over the observed period, which can be attributed to improvements in ridge--valley delineation as illustrated in Figure~\ref{fig:intro-ageing}: poorly defined valleys in younger subjects artificially widen ridges, reducing the measured frequency.

However, progression is not strictly harmonious across all methods. At 15 days, dominant frequency estimates already align with those of older age groups, while at 1 month the lower ridge-width bound decreases. Similar fluctuations across early infant age groups have been reported in prior work \cite{Camacho2017, Preciozzi2020}. Furthermore, the ridge-to-valley width ratios inferred for newborns are consistent with \cite{Saggese2019}, where ridges were approximately four times larger than valleys, and the valley-width intervals for the 1 to 3 month groups align with the averages reported by \cite{Camacho2017, Preciozzi2020}, noting that direct numerical correspondence is not expected as the present framework operates on integer-valued pixel intervals.

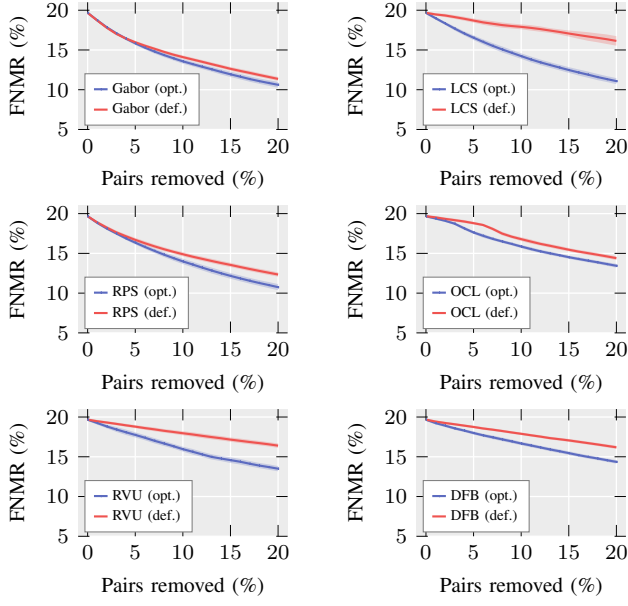
\begin{figure}[!htb]
    \centering
    \begin{subfigure}[b]{0.24\textwidth}
        \centering
        \pgfplotstableread{
x	y       std
0	19.67	0.17
1	18.74	0.16
2	17.86	0.17
3	17.11	0.16
4	16.44	0.20
5	15.84	0.22
6	15.30	0.23
7	14.81	0.26
8	14.37	0.27
9	13.96	0.28
10	13.57	0.29
11	13.23	0.28
12	12.90	0.29
13	12.58	0.30
14	12.25	0.33
15	11.94	0.34
16	11.65	0.34
17	11.35	0.34
18	11.09	0.34
19	10.85	0.35
20	10.62	0.37
}\dataA

\pgfplotstableread{
x    y      std
0	19.67	0.17
1	18.71	0.16
2	17.80	0.17
3	17.03	0.17
4	16.45	0.17
5	15.97	0.19
6	15.54	0.20
7	15.16	0.19
8	14.77	0.19
9	14.41	0.20
10	14.10	0.22
11	13.81	0.22
12	13.52	0.22
13	13.22	0.23
14	12.93	0.23
15	12.63	0.24
16	12.37	0.25
17	12.12	0.26
18	11.86	0.25
19	11.61	0.25
20	11.36	0.25
}\dataB

\begin{tikzpicture}[baseline]
\begin{axis}[
    width=\columnwidth,
    height=0.75\columnwidth,
    xmin=-1, xmax=21,
    ymin=5, ymax=21,
    xtick={0, 5, 10, 15, 20},
    ytick={0, 5, 10, 15, 20},
    xlabel=Pairs removed (\%),
    ylabel=FNMR (\%),
    tick label style={font=\footnotesize},
    label style={font=\footnotesize},
    axis background/.style={fill=gray!15},
    grid=both,
    grid style={white, line width=0.5pt},
    axis line style={draw=none},
    tick style={draw=black},
    clip=false,
    legend pos=south west,
    legend style={font=\footnotesize, fill=white, draw=gray, nodes={scale=0.7, transform shape}},
    legend image post style={xscale=0.3},
]

\addplot[name path=Aup, draw=none, forget plot]
  table[x=x, y expr=\thisrow{y}+\thisrow{std}] {\dataA};
\addplot[name path=Alo, draw=none, forget plot]
  table[x=x, y expr=\thisrow{y}-\thisrow{std}] {\dataA};
\addplot[matblue, opacity=0.25, forget plot] fill between[of=Aup and Alo];

\addplot[matblue, thick, line join=round, mark size=0pt, mark=x]
  table[x=x, y=y] {\dataA};
\addlegendentry{Gabor (opt.)}

\addplot[name path=Bup, draw=none, forget plot]
  table[x=x, y expr=\thisrow{y}+\thisrow{std}] {\dataB};
\addplot[name path=Blo, draw=none, forget plot]
  table[x=x, y expr=\thisrow{y}-\thisrow{std}] {\dataB};
\addplot[matred, opacity=0.25, forget plot] fill between[of=Bup and Blo];

\addplot[matred, thick, line join=round, mark size=0pt, mark=*]
  table[x=x, y=y] {\dataB};
\addlegendentry{Gabor (def.)}

\end{axis}
\end{tikzpicture}
    \end{subfigure}
    \begin{subfigure}[b]{0.24\textwidth}
        \centering
        \pgfplotstableread{
x    y      std
0	19.67	0.17
1	19.00	0.19
2	18.35	0.19
3	17.70	0.25
4	17.09	0.26
5	16.54	0.30
6	16.02	0.32
7	15.52	0.35
8	15.07	0.37
9	14.65	0.36
10	14.23	0.35
11	13.82	0.36
12	13.45	0.35
13	13.12	0.35
14	12.80	0.35
15	12.48	0.36
16	12.18	0.38
17	11.90	0.39
18	11.60	0.39
19	11.34	0.37
20	11.08	0.38
}\dataA

\pgfplotstableread{
x    y      std
0	19.67	0.17
1	19.47	0.16
2	19.34	0.21
3	19.14	0.23
4	18.93	0.25
5	18.70	0.26
6	18.47	0.30
7	18.30	0.33
8	18.13	0.33
9	18.01	0.31
10	17.91	0.31
11	17.78	0.34
12	17.63	0.34
13	17.45	0.30
14	17.27	0.33
15	17.07	0.38
16	16.88	0.47
17	16.70	0.54
18	16.53	0.56
19	16.35	0.60
20	16.14	0.60
}\dataB

\begin{tikzpicture}[baseline]
\begin{axis}[
    width=\columnwidth,
    height=0.75\columnwidth,
    xmin=-1, xmax=21,
    ymin=5, ymax=21,
    xtick={0, 5, 10, 15, 20},
    ytick={0, 5, 10, 15, 20},
    xlabel=Pairs removed (\%),
    ylabel=FNMR (\%),
    tick label style={font=\footnotesize},
    label style={font=\footnotesize},
    axis background/.style={fill=gray!15},
    grid=both,
    grid style={white, line width=0.5pt},
    axis line style={draw=none},
    tick style={draw=black},
    clip=false,
    legend pos=south west,
    legend style={font=\footnotesize, fill=white, draw=gray, nodes={scale=0.7, transform shape}},
    legend image post style={xscale=0.3},
]

\addplot[name path=Aup, draw=none, forget plot]
  table[x=x, y expr=\thisrow{y}+\thisrow{std}] {\dataA};
\addplot[name path=Alo, draw=none, forget plot]
  table[x=x, y expr=\thisrow{y}-\thisrow{std}] {\dataA};
\addplot[matblue, opacity=0.25, forget plot] fill between[of=Aup and Alo];

\addplot[matblue, thick, line join=round, mark size=0pt, mark=x]
  table[x=x, y=y] {\dataA};
\addlegendentry{LCS (opt.)}

\addplot[name path=Bup, draw=none, forget plot]
  table[x=x, y expr=\thisrow{y}+\thisrow{std}] {\dataB};
\addplot[name path=Blo, draw=none, forget plot]
  table[x=x, y expr=\thisrow{y}-\thisrow{std}] {\dataB};
\addplot[matred, opacity=0.25, forget plot] fill between[of=Bup and Blo];

\addplot[matred, thick, line join=round, mark size=0pt, mark=*]
  table[x=x, y=y] {\dataB};
\addlegendentry{LCS (def.)}

\end{axis}
\end{tikzpicture}
    \end{subfigure}
    \begin{subfigure}[b]{0.24\textwidth}
        \centering
        \pgfplotstableread{
x	y       std
0	19.67	0.17
1	18.84	0.17
2	18.13	0.18
3	17.51	0.18
4	16.90	0.20
5	16.35	0.23
6	15.82	0.26
7	15.32	0.28
8	14.85	0.30
9	14.41	0.33
10	13.99	0.34
11	13.61	0.35
12	13.22	0.35
13	12.85	0.37
14	12.50	0.36
15	12.17	0.36
16	11.86	0.36
17	11.56	0.36
18	11.28	0.36
19	11.02	0.38
20	10.75	0.39

}\dataA

\pgfplotstableread{
x    y      std
0	19.67	0.17
1	18.90	0.19
2	18.25	0.21
3	17.65	0.22
4	17.16	0.23
5	16.70	0.25
6	16.28	0.26
7	15.89	0.25
8	15.53	0.25
9	15.20	0.24
10	14.89	0.27
11	14.59	0.26
12	14.32	0.27
13	14.05	0.27
14	13.79	0.28
15	13.55	0.27
16	13.29	0.26
17	13.04	0.27
18	12.80	0.27
19	12.57	0.28
20	12.34	0.27
}\dataB

\begin{tikzpicture}[baseline]
\begin{axis}[
    width=\columnwidth,
    height=0.75\columnwidth,
    xmin=-1, xmax=21,
    ymin=5, ymax=21,
    xtick={0, 5, 10, 15, 20},
    ytick={0, 5, 10, 15, 20},
    xlabel=Pairs removed (\%),
    ylabel=FNMR (\%),
    tick label style={font=\footnotesize},
    label style={font=\footnotesize},
    axis background/.style={fill=gray!15},
    grid=both,
    grid style={white, line width=0.5pt},
    axis line style={draw=none},
    tick style={draw=black},
    clip=false,
    legend pos=south west,
    legend style={font=\footnotesize, fill=white, draw=gray, nodes={scale=0.7, transform shape}},
    legend image post style={xscale=0.3},
]

\addplot[name path=Aup, draw=none, forget plot]
  table[x=x, y expr=\thisrow{y}+\thisrow{std}] {\dataA};
\addplot[name path=Alo, draw=none, forget plot]
  table[x=x, y expr=\thisrow{y}-\thisrow{std}] {\dataA};
\addplot[matblue, opacity=0.25, forget plot] fill between[of=Aup and Alo];

\addplot[matblue, thick, line join=round, mark size=0pt, mark=x]
  table[x=x, y=y] {\dataA};
\addlegendentry{RPS (opt.)}

\addplot[name path=Bup, draw=none, forget plot]
  table[x=x, y expr=\thisrow{y}+\thisrow{std}] {\dataB};
\addplot[name path=Blo, draw=none, forget plot]
  table[x=x, y expr=\thisrow{y}-\thisrow{std}] {\dataB};
\addplot[matred, opacity=0.25, forget plot] fill between[of=Bup and Blo];

\addplot[matred, thick, line join=round, mark size=0pt, mark=*]
  table[x=x, y=y] {\dataB};
\addlegendentry{RPS (def.)}

\end{axis}
\end{tikzpicture}
    \end{subfigure}
    \begin{subfigure}[b]{0.24\textwidth}
        \centering
        \pgfplotstableread{
x	y       std
0	19.67	0.17
1	19.39	0.17
2	19.10	0.15
3	18.74	0.17
4	18.16	0.21
5	17.63	0.19
6	17.21	0.24
7	16.83	0.20
8	16.50	0.20
9	16.19	0.19
10	15.86	0.21
11	15.55	0.22
12	15.26	0.22
13	15.01	0.22
14	14.77	0.21
15	14.52	0.21
16	14.30	0.21
17	14.09	0.22
18	13.88	0.21
19	13.66	0.22
20	13.44	0.24
}\dataA

\pgfplotstableread{
x    y      std
0	19.67	0.17
1	19.53	0.17
2	19.34	0.18
3	19.18	0.19
4	19.01	0.19
5	18.80	0.18
6	18.57	0.16
7	18.08	0.20
8	17.49	0.23
9	17.10	0.21
10	16.79	0.24
11	16.49	0.26
12	16.20	0.22
13	15.96	0.23
14	15.71	0.24
15	15.46	0.22
16	15.23	0.22
17	15.03	0.23
18	14.83	0.23
19	14.62	0.24
20	14.40	0.24
}\dataB

\begin{tikzpicture}[baseline]
\begin{axis}[
    width=\columnwidth,
    height=0.75\columnwidth,
    xmin=-1, xmax=21,
    ymin=5, ymax=21,
    xtick={0, 5, 10, 15, 20},
    ytick={0, 5, 10, 15, 20},
    xlabel=Pairs removed (\%),
    ylabel=FNMR (\%),
    tick label style={font=\footnotesize},
    label style={font=\footnotesize},
    axis background/.style={fill=gray!15},
    grid=both,
    grid style={white, line width=0.5pt},
    axis line style={draw=none},
    tick style={draw=black},
    clip=false,
    legend pos=south west,
    legend style={font=\footnotesize, fill=white, draw=gray, nodes={scale=0.7, transform shape}},
    legend image post style={xscale=0.3},
]

\addplot[name path=Aup, draw=none, forget plot]
  table[x=x, y expr=\thisrow{y}+\thisrow{std}] {\dataA};
\addplot[name path=Alo, draw=none, forget plot]
  table[x=x, y expr=\thisrow{y}-\thisrow{std}] {\dataA};
\addplot[matblue, opacity=0.25, forget plot] fill between[of=Aup and Alo];

\addplot[matblue, thick, line join=round, mark size=0pt, mark=x]
  table[x=x, y=y] {\dataA};
\addlegendentry{OCL (opt.)}

\addplot[name path=Bup, draw=none, forget plot]
  table[x=x, y expr=\thisrow{y}+\thisrow{std}] {\dataB};
\addplot[name path=Blo, draw=none, forget plot]
  table[x=x, y expr=\thisrow{y}-\thisrow{std}] {\dataB};
\addplot[matred, opacity=0.25, forget plot] fill between[of=Bup and Blo];

\addplot[matred, thick, line join=round, mark size=0pt, mark=*]
  table[x=x, y=y] {\dataB};
\addlegendentry{OCL (def.)}

\end{axis}
\end{tikzpicture}
    \end{subfigure}
    \begin{subfigure}[b]{0.24\textwidth}
        \centering
        \pgfplotstableread{
x	y       std
0	19.67	0.17
1	19.25	0.21
2	18.82	0.24
3	18.44	0.31
4	18.09	0.34
5	17.75	0.33
6	17.40	0.32
7	17.02	0.33
8	16.68	0.32
9	16.33	0.30
10	15.96	0.31
11	15.64	0.35
12	15.31	0.33
13	14.99	0.33
14	14.78	0.32
15	14.58	0.36
16	14.38	0.35
17	14.13	0.34
18	13.90	0.33
19	13.71	0.34
20	13.50	0.33
}\dataA

\pgfplotstableread{
x    y      std
0	19.67	0.17
1	19.47	0.17
2	19.31	0.17
3	19.14	0.18
4	18.96	0.19
5	18.79	0.20
6	18.61	0.22
7	18.45	0.24
8	18.27	0.23
9	18.12	0.24
10	17.96	0.26
11	17.81	0.27
12	17.64	0.28
13	17.48	0.28
14	17.33	0.28
15	17.16	0.27
16	17.01	0.28
17	16.86	0.28
18	16.72	0.28
19	16.57	0.27
20	16.40	0.28
}\dataB

\begin{tikzpicture}[baseline]
\begin{axis}[
    width=\columnwidth,
    height=0.75\columnwidth,
    xmin=-1, xmax=21,
    ymin=5, ymax=21,
    xtick={0, 5, 10, 15, 20},
    ytick={0, 5, 10, 15, 20},
    xlabel=Pairs removed (\%),
    ylabel=FNMR (\%),
    tick label style={font=\footnotesize},
    label style={font=\footnotesize},
    axis background/.style={fill=gray!15},
    grid=both,
    grid style={white, line width=0.5pt},
    axis line style={draw=none},
    tick style={draw=black},
    clip=false,
    legend pos=south west,
    legend style={font=\footnotesize, fill=white, draw=gray, nodes={scale=0.7, transform shape}},
    legend image post style={xscale=0.3},
]

\addplot[name path=Aup, draw=none, forget plot]
  table[x=x, y expr=\thisrow{y}+\thisrow{std}] {\dataA};
\addplot[name path=Alo, draw=none, forget plot]
  table[x=x, y expr=\thisrow{y}-\thisrow{std}] {\dataA};
\addplot[matblue, opacity=0.25, forget plot] fill between[of=Aup and Alo];

\addplot[matblue, thick, line join=round, mark size=0pt, mark=x]
  table[x=x, y=y] {\dataA};
\addlegendentry{RVU (opt.)}

\addplot[name path=Bup, draw=none, forget plot]
  table[x=x, y expr=\thisrow{y}+\thisrow{std}] {\dataB};
\addplot[name path=Blo, draw=none, forget plot]
  table[x=x, y expr=\thisrow{y}-\thisrow{std}] {\dataB};
\addplot[matred, opacity=0.25, forget plot] fill between[of=Bup and Blo];

\addplot[matred, thick, line join=round, mark size=0pt, mark=*]
  table[x=x, y=y] {\dataB};
\addlegendentry{RVU (def.)}

\end{axis}
\end{tikzpicture}
    \end{subfigure}
    \begin{subfigure}[b]{0.24\textwidth}
        \centering
        \pgfplotstableread{
x	y       std
0	19.67	0.17
1	19.24	0.17
2	18.92	0.19
3	18.58	0.19
4	18.30	0.19
5	17.99	0.20
6	17.70	0.20
7	17.45	0.23
8	17.19	0.23
9	16.94	0.23
10	16.67	0.22
11	16.43	0.23
12	16.18	0.21
13	15.93	0.23
14	15.70	0.23
15	15.46	0.23
16	15.21	0.23
17	14.99	0.21
18	14.79	0.22
19	14.57	0.22
20	14.36	0.22
}\dataA

\pgfplotstableread{
x    y      std
0	19.67	0.17
1	19.43	0.16
2	19.25	0.16
3	19.09	0.16
4	18.92	0.16
5	18.75	0.17
6	18.56	0.17
7	18.41	0.17
8	18.24	0.17
9	18.06	0.17
10	17.89	0.17
11	17.71	0.17
12	17.54	0.18
13	17.35	0.18
14	17.21	0.18
15	17.06	0.19
16	16.89	0.19
17	16.72	0.19
18	16.55	0.19
19	16.37	0.20
20	16.19	0.20
}\dataB

\begin{tikzpicture}[baseline]
\begin{axis}[
    width=\columnwidth,
    height=0.75\columnwidth,
    xmin=-1, xmax=21,
    ymin=5, ymax=21,
    xtick={0, 5, 10, 15, 20},
    ytick={0, 5, 10, 15, 20},
    xlabel=Pairs removed (\%),
    ylabel=FNMR (\%),
    tick label style={font=\footnotesize},
    label style={font=\footnotesize},
    axis background/.style={fill=gray!15},
    grid=both,
    grid style={white, line width=0.5pt},
    axis line style={draw=none},
    tick style={draw=black},
    clip=false,
    legend pos=south west,
    legend style={font=\footnotesize, fill=white, draw=gray, nodes={scale=0.7, transform shape}},
    legend image post style={xscale=0.3},
]

\addplot[name path=Aup, draw=none, forget plot]
  table[x=x, y expr=\thisrow{y}+\thisrow{std}] {\dataA};
\addplot[name path=Alo, draw=none, forget plot]
  table[x=x, y expr=\thisrow{y}-\thisrow{std}] {\dataA};
\addplot[matblue, opacity=0.25, forget plot] fill between[of=Aup and Alo];

\addplot[matblue, thick, line join=round, mark size=0pt, mark=x]
  table[x=x, y=y] {\dataA};
\addlegendentry{DFB (opt.)}

\addplot[name path=Bup, draw=none, forget plot]
  table[x=x, y expr=\thisrow{y}+\thisrow{std}] {\dataB};
\addplot[name path=Blo, draw=none, forget plot]
  table[x=x, y expr=\thisrow{y}-\thisrow{std}] {\dataB};
\addplot[matred, opacity=0.25, forget plot] fill between[of=Bup and Blo];

\addplot[matred, thick, line join=round, mark size=0pt, mark=*]
  table[x=x, y=y] {\dataB};
\addlegendentry{DFB (def.)}

\end{axis}
\end{tikzpicture}
    \end{subfigure}
    \caption{FNMR plots for default (def.) and tuned (opt.) parameters for all methods (one per panel), accumulated across all six age groups and all 30 folds. Lower values indicate better performance.}
    \label{fig:bayes-fnmr}
\end{figure}

The resulting effect of parameter tuning in the cross-validation experiment is an accelerated decay of FNMR for the tuned (opt.) configurations compared to the default (def.) values. Figure~\ref{fig:bayes-fnmr} presents the FNMR obtained on the held-out subjects of each cross-validation fold, using parameters optimized exclusively on the corresponding training folds. The results show that tuned configurations consistently reduce FNMR relative to the default settings, although the magnitude of the improvement varies across methods. In particular, LCS is more responsive to parameter tuning, whereas Gabor and OCL achieve only minor performance gains.
 
\begin{figure}[!htb]
    \centering
    \input{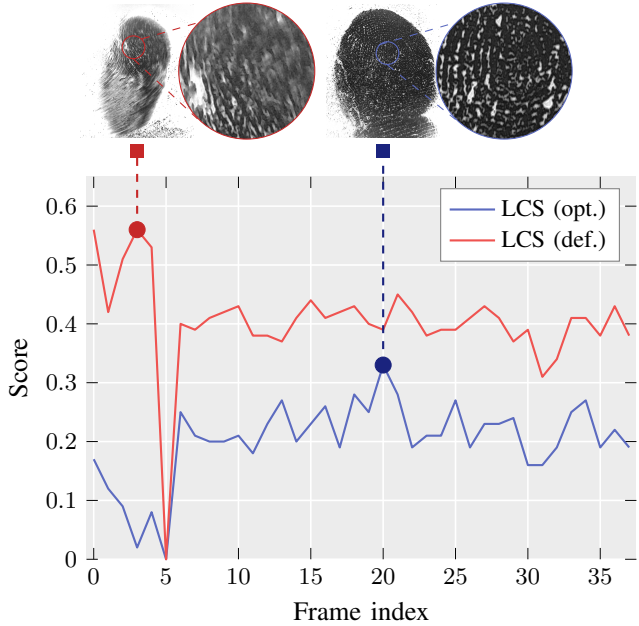}
    \caption{LCS scoring of a newborn thumb (test only subset). The fingerprint core region is showcased for the best frames selected by the default and tuned configurations.}
    \label{fig:lcs-score}
\end{figure}

Figure~\ref{fig:lcs-score} illustrates how the different default (def.) and tuned (opt.) scores behave over videos using LCS and a newborn thumb captured hours after birth. Tuned LCS assigns its highest score to a more informative frame, whereas the default configuration selects a blurred mid-movement frame. The key difference in gallery selection is that tuning produces more discriminative quality scores because the parameter bounds are adapted to the feature distribution of newborn fingerprints, thereby enabling more effective filtering.

\subsection{Recognition Results}

To evaluate the performance of the quality assessment methods across age group subsets, each method independently constructs a 5-frame gallery for every digit, selecting frames according to its own quality score ranking while enforcing a minimum inter-frame distance of three frames, as described in Section~\ref{subsec:met-data}. Similar to the FVC protocol \cite{Maio2002}, each identity then contributes a single sample (highest average genuine comparison score) to compose the gallery of impostors, which is fixed per age group and invariant across methods to ensure that differences in performance reflect only the genuine gallery composition. This is made possible because genuine comparisons are restricted to frames within the same video. Accordingly, each frame is subject to 4 genuine and 89 impostor comparisons, totaling 1,800 genuine and 40,050 impostor pairs per method/age group and fold. NFIQ2 is used as the reference baseline, being the most consolidated adult quality method; the default configurations of the remaining methods provide a second baseline.

Table~\ref{tab:res-tf-utfpr} presents VeriFinger 2025.2 TMR@FMR=0.1\% and 1.0\% results for each quality estimation method across all age groups in our test subset of UTFPR-IFD. An accept is defined as the successful matching of individual frames to enable a detailed evaluation of gallery utility, though higher rates can be achieved with stricter protocols as Section~\ref{sec:res-fusion} later explores.

\begin{table*}[htbp]
\centering
\caption{Mean matching performance (TMR@FMR=0.1\% and $1.0\%$) by age and method (90 digits per fold, 5 folds) with standard deviation. Bold values indicate best results per method, underlined values indicate best results for each age.}
\label{tab:res-tf-utfpr}

\scriptsize
\setlength{\tabcolsep}{4pt}
\renewcommand{\arraystretch}{1.1}
\newcommand{\sd}[1]{{\tiny$\pm$#1}}

\begin{tabular}{l l 
>{\columncolor{gray!15}}c >{\columncolor{gray!15}}c
c c
>{\columncolor{gray!15}}c >{\columncolor{gray!15}}c
c c
>{\columncolor{gray!15}}c >{\columncolor{gray!15}}c
c c}
\toprule
\multirow{2}{*}{Method} & \multirow{2}{*}{Setting} 
& \multicolumn{2}{c}{Newborn} 
& \multicolumn{2}{c}{7-day} 
& \multicolumn{2}{c}{15-day} 
& \multicolumn{2}{c}{1-month} 
& \multicolumn{2}{c}{2-month} 
& \multicolumn{2}{c}{3-month} \\
\cmidrule(lr){3-4} \cmidrule(lr){5-6} \cmidrule(lr){7-8}
\cmidrule(lr){9-10} \cmidrule(lr){11-12} \cmidrule(lr){13-14}
& & 0.1\% & 1.0\% & 0.1\% & 1.0\% & 0.1\% & 1.0\% & 0.1\% & 1.0\% & 0.1\% & 1.0\% & 0.1\% & 1.0\% \\
\midrule

\multirow{2}{*}{Gabor}
& tuned & \textbf{30.6} \sd{6.6} & \textbf{46.8} \sd{6.5} & 53.3 \sd{5.5} & \textbf{75.8} \sd{4.2} & \textbf{70.7} \sd{6.7} & \textbf{86.3} \sd{3.8} & 77.9 \sd{6.4} & \textbf{87.3} \sd{4.5} & \textbf{80.2} \sd{5.3} & \textbf{88.4} \sd{2.4} & 72.7 \sd{7.5} & \textbf{84.7} \sd{4.8} \\
& default   & 29.4 \sd{6.7} & 44.2 \sd{8.3} & \textbf{54.1} \sd{8.1} & 73.8 \sd{5.9} & 68.2 \sd{7.4} & 83.0 \sd{4.9} & \textbf{78.4} \sd{6.9} & 87.0 \sd{4.9} & 80.0 \sd{6.2} & 88.1 \sd{3.6} & \textbf{73.6} \sd{6.5} & 84.2 \sd{4.4} \\  \cmidrule(lr){2-14}

\multirow{2}{*}{LCS}
& tuned & \textbf{34.5} \sd{5.7} & \underline{\textbf{53.6}} \sd{6.9} & \textbf{57.1} \sd{7.2} & \underline{\textbf{78.7}} \sd{5.4} & \textbf{73.1} \sd{4.5} & \underline{\textbf{88.0}} \sd{3.1} & 75.8 \sd{7.4} & \textbf{86.7} \sd{5.0} & \textbf{79.9} \sd{5.6} & \textbf{87.3} \sd{3.6} & \textbf{71.8} \sd{10.0} & \textbf{82.8} \sd{6.3} \\
& default   & 27.5 \sd{4.9} & 45.3 \sd{6.3} & 50.5 \sd{7.1} & 72.1 \sd{4.9} & 65.3 \sd{5.5} & 82.0 \sd{4.1} & \textbf{76.5} \sd{6.6} & 84.9 \sd{4.9} & 75.8 \sd{6.3} & 83.8 \sd{5.3} & 68.1 \sd{8.7} & 80.3 \sd{6.6} \\
\cmidrule(lr){2-14}

\multirow{2}{*}{RPS}
& tuned & \textbf{30.2} \sd{6.0} & \textbf{47.5} \sd{6.2} & \textbf{53.8} \sd{7.2} & \textbf{77.3} \sd{5.0} & \underline{\textbf{73.8}} \sd{7.3} & \textbf{87.3} \sd{4.1} & \textbf{80.1} \sd{6.6} & \textbf{88.4} \sd{3.7} & \underline{\textbf{83.2}} \sd{6.2} & \underline{\textbf{90.5}} \sd{2.9} & \underline{\textbf{78.4}} \sd{6.3} & \underline{\textbf{88.8}} \sd{3.7} \\
& default   & 28.9 \sd{6.8} & 44.4 \sd{6.5} & 53.1 \sd{6.8} & 74.1 \sd{6.4} & 66.9 \sd{7.8} & 82.7 \sd{5.7} & 76.6 \sd{7.8} & 85.5 \sd{4.9} & 76.9 \sd{6.6} & 85.4 \sd{5.0} & 71.5 \sd{6.4} & 83.0 \sd{4.5} \\

\midrule

\multirow{2}{*}{OCL}
& tuned & \textbf{24.1} \sd{6.4} & \textbf{38.0} \sd{9.3} & \textbf{53.9} \sd{7.9} & \textbf{72.5} \sd{5.6} & \textbf{67.7} \sd{6.0} & \textbf{81.9} \sd{4.0} & \textbf{73.7} \sd{4.8} & \textbf{83.5} \sd{2.2} & \textbf{74.2} \sd{8.3} & \textbf{82.5} \sd{6.6} & \textbf{62.7} \sd{9.3} & \textbf{73.0} \sd{8.1} \\
& default   & 23.2 \sd{6.6} & 35.7 \sd{8.6} & 53.6 \sd{8.1} & 71.9 \sd{5.4} & 65.6 \sd{7.0} & 80.9 \sd{4.6} & 72.4 \sd{5.6} & 82.0 \sd{2.4} & 73.1 \sd{9.1} & 81.8 \sd{6.9} & 61.9 \sd{10.9} & 71.9 \sd{9.1} \\
\cmidrule(lr){2-14}

\multirow{2}{*}{RVU}
& tuned & \textbf{26.2} \sd{7.2} & \textbf{41.9} \sd{8.4} & \textbf{50.8} \sd{8.7} & \textbf{71.5} \sd{5.9} & \textbf{65.5} \sd{5.5} & \textbf{82.1} \sd{3.6} & \textbf{73.9} \sd{5.4} & \textbf{83.5} \sd{2.8} & \textbf{74.5} \sd{9.1} & \textbf{83.3} \sd{5.9} & \textbf{69.2} \sd{7.1} & \textbf{81.4} \sd{4.4} \\
& default   & 20.1 \sd{7.1} & 33.2 \sd{9.5} & 47.9 \sd{7.8} & 68.3 \sd{5.7} & 59.9 \sd{7.0} & 76.8 \sd{4.5} & 71.2 \sd{5.7} & 79.7 \sd{2.9} & 70.5 \sd{8.7} & 78.0 \sd{6.8} & 60.5 \sd{10.5} & 70.0 \sd{9.7} \\
\cmidrule(lr){2-14}

\multirow{2}{*}{DFB}
& tuned & \textbf{29.6} \sd{11.1} & \textbf{44.3} \sd{14.1} & \textbf{53.8} \sd{6.2} & \textbf{76.1} \sd{5.2} & \textbf{68.7} \sd{5.2} & \textbf{85.5} \sd{4.2} & \textbf{75.5} \sd{6.8} & \textbf{85.4} \sd{4.4} & \textbf{78.7} \sd{6.3} & \textbf{86.9} \sd{4.7} & \textbf{67.7} \sd{11.4} & \textbf{79.0} \sd{9.0} \\
& default   & 24.3 \sd{9.0} & 37.4 \sd{14.2} & 52.8 \sd{6.4} & 72.6 \sd{6.4} & 67.6 \sd{5.4} & 82.0 \sd{3.8} & 73.0 \sd{4.3} & 81.7 \sd{2.1} & 71.3 \sd{9.5} & 79.8 \sd{6.5} & 62.2 \sd{11.1} & 71.6 \sd{9.1} \\

\midrule

NFIQ2 & --- & 26.9 \sd{6.9} & 42.5 \sd{8.6} & 52.3 \sd{6.6} & 74.1 \sd{5.9} & 68.8 \sd{6.1} & 84.5 \sd{5.1} & 77.8 \sd{5.1} & 85.7 \sd{3.2} & 76.9 \sd{7.3} & 85.6 \sd{5.2} & 68.9 \sd{9.1} & 80.5 \sd{5.9} \\
Mean & --- & 27.7 \sd{5.2} & 42.2 \sd{6.5} & 49.6 \sd{7.5} & 69.4 \sd{6.4} & 62.2 \sd{8.0} & 77.4 \sd{5.3} & 71.6 \sd{6.9} & 80.5 \sd{4.1} & 73.5 \sd{7.8} & 81.1 \sd{5.3} & 69.1 \sd{8.5} & 78.6 \sd{6.6} \\
SIG & --- & 27.7 \sd{7.3} & 45.9 \sd{9.5} & \underline{\textbf{59.1}} \sd{7.8} & 78.0 \sd{4.8} & 72.4 \sd{4.3} & 86.8 \sd{2.2} & 79.1 \sd{5.6} & 87.5 \sd{3.5} & 78.0 \sd{6.5} & 86.6 \sd{4.5} & 70.8 \sd{9.6} & 81.8 \sd{6.9} \\
MQCDA & --- & \underline{\textbf{35.0}} \sd{4.6} & 51.2 \sd{4.5} & 57.6 \sd{8.8} & 76.2 \sd{6.5} & 72.5 \sd{5.5} & 86.5 \sd{4.3} & \underline{\textbf{80.5}} \sd{6.9} & \underline{\textbf{88.5}} \sd{3.4} & 80.9 \sd{6.1} & 88.2 \sd{3.9} & 76.5 \sd{7.8} & 85.8 \sd{4.8} \\

\bottomrule
\end{tabular}
\end{table*}

A complementary Friedman analysis, followed by the post-hoc Nemenyi test~\cite{Demvsar2006}, was conducted to compare the quality assessment methods across age groups and folds ($6$ age groups $\times$ $5$ folds $= 30$ blocks). Table~\ref{tab:res-friedman} reports the mean ranks at FMR=1.0\%, adopted as the reference operating point because single-frame comparisons are inherently more error-prone than the decisions of biometric systems, which can aggregate several fingers. Two methods are considered significantly different only if their mean ranks differ by more than the critical difference ($\mathrm{CD}=4.2$, $\alpha=0.05$).

\begin{table}[!htbp]
\centering
\scriptsize
\caption{Friedman-Nemenyi ranking analysis of quality methods (lower rank = better performance) at FMR=1.0\%.}
\label{tab:res-friedman}
\setlength{\tabcolsep}{15pt}
\begin{tabular}{llS[table-format=2.2]S[table-format=2.2]}
\toprule
\textbf{Method} & \textbf{Setting} & \textbf{Mean Rank} \\
\midrule
RPS    & tuned   &  2.8 \\
MQCDA  & ---     &  3.6 \\
LCS    & tuned   &  4.4 \\
Gabor  & tuned   &  4.6 \\
SIG    & ---     &  4.9 \\
DFB    & tuned   &  6.8 \\
Gabor  & default &  6.9 \\
RPS    & default &  7.8 \\
NFIQ2  & ---     &  8.3 \\
LCS    & default &  10.0 \\
RVU    & tuned   &  10.3 \\
OCL    & tuned   &  11.7 \\
DFB    & default &  12.1 \\
MU     & ---     &  13.0 \\
OCL    & default &  13.2 \\
RVU    & default &  15.5 \\
\bottomrule
\end{tabular}
\begin{minipage}{0.45\textwidth}
\small
\vspace{4pt}
CD $=4.2$ ($\alpha=0.05$ \textit{Nemenyi test}), $\chi^2_F(15) = 300.85$, $p < 0.001$.
\end{minipage}
\end{table}

Tuned configurations obtained higher mean TMR than their defaults in every method-age combination (36/36), at FMR=1.0\%, and in 32 of 36 at FMR=0.1\% (Table~\ref{tab:res-tf-utfpr}). The exceptions at 0.1\% were Gabor (three age groups), and LCS (1-month). Out of all the evaluated methods, tuned RPS obtained the lowest mean rank ($2.8$), although it was not significantly different from the other top-five methods of Table~\ref{tab:res-friedman}. It obtained the highest mean TMR for the 2- and 3-month groups at both FMRs, and for the 15-day group at FMR=0.1\%, but was less competitive for newborns ($47.5\%$ vs. $53.6\%$ for LCS at FMR=1.0\%). Tuning contributed to this outcome: tuned RPS ranks $5.0$ units better than its default configuration at FMR=1.0\%, exceeding the CD, and its TMR gains at FMR=0.1\% range from $0.7$ to $6.9$ points (tuned vs. default).

Tuned LCS also improved significantly over its default configuration, with a mean-rank difference of $\Delta=5.6$, above the CD. This is consistent with expectations regarding LCS's dependence on ridge and valley width ranges, which differ between infants and the adult standard and vary with age. In Table~\ref{tab:res-tf-utfpr}, tuned LCS obtained the highest mean TMR at FMR=1.0\% for infants aged 15 days or younger, although its margins over the next best method ($2.4$, $0.7$ and $0.7$ points for newborn, 7-day and 15-day) are within the fold standard deviation. At FMR=0.1\%, its gains over the default configuration reached $7.8$ points (15-day), with the 1-month group as the only exception ($-0.7$). It also obtained the third lowest mean rank at FMR=1.0\% ($4.4$).

Gabor tuning, however, presented mixed results. It did not obtain the highest mean TMR in any age group of Table~\ref{tab:res-tf-utfpr}, and tuning did not consistently improve on the default setting at FMR=0.1\%: the default configuration achieved higher mean TMR for the 7-day, 1-month and 3-month groups, with differences of at most $0.9$ point. At FMR=1.0\%, tuned Gabor obtained higher mean TMR in all six age groups (by $0.3$ to $3.3$ points). This was sufficient to place it among the top five methods at FMR=1.0\% in Table~\ref{tab:res-friedman}, however the mean rank difference of $\Delta=2.3$ remained below the CD.

MQCDA and SIG, which underwent no optimization, complete the top five rank of Table~\ref{tab:res-friedman}. MQCDA is a commercial deep-learning model whose architecture and training data are not publicly documented, and it is the only method evaluated that specifically targets children, which may explain its competitive performance across age groups. At FMR=0.1\%, it obtained the highest mean TMR for newborns ($35.0\%$) and 1-month infants ($80.5\%$), with margins of $0.5$ and $0.4$ points over the next best method. SIG, the standard deviation of pixel gray-level intensities, is independent of age and resolution and makes no assumptions about fingerprint-specific traits, making it a plausible alternative for datasets with variable characteristics such as evolving infant fingerprints. It obtained the highest mean TMR for the 7-day group at FMR=0.1\% ($59.1\%$, $1.5$ points over MQCDA).

Within the top five methods, no pair differs significantly at FMR=1.0\% (Table~\ref{tab:res-friedman}), as the largest rank gap ($2.1$) is below the CD ($4.2$). The remaining methods produced mixed results. 

Tuned DFB performed significantly better than its default setting ($\Delta=5.3$), obtaining higher mean TMR than its default in all six age groups at FMR=0.1\%, and ranked above NFIQ2 ($6.8$ vs. $8.3$), although not significantly. Considering its performance against standard methods, it may represent a strong filter-based alternative to more traditional quality measures. Block-size tuning of OCL and RVU also improved their results, but not sufficiently to improve rankings above the NFIQ2 baseline. For RVU the rank gain was significant ($\Delta=5.2$), which translated to TMR gains ranging from $2.7$ to $11.4$ points at both FMR thresholds. For OCL results were modest, with gains to TMR means capped at $2.1$ points and a mean rank difference of $1.5$, below the CD.

Lastly, NFIQ2 ranked above its constituent features with the notable exception of SIG and tuned LCS. This difference was significant for MU, and default OCL and RVU (rank gaps of $4.7$, $4.9$ and $7.2$, respectively). However, incorporating tuned versions of these features into an infant focused index could still improve its performance on infant fingerprints. In particularly LCS, which obtained higher mean TMR than NFIQ2 in all six age groups at FMR=1.0\% (five of six at FMR=0.1\%), although its rank gap to NFIQ2 at $\Delta=3.9$ was slightly below the CD. In addition to current NFIQ2 features, overall recognition results suggest that a composite quality measure designed specifically for children might benefit from combining additional methods such as RPS and Gabor.

\begin{figure}[!htb]
\centering
\begin{tikzpicture}[baseline]
\begin{semilogxaxis}[
    width=\columnwidth,
    height=0.75\columnwidth,
    xmin=0.1, xmax=100,
    ymin=20, ymax=100,
    xtick={0.1,1,10,100},
    xticklabels={0.1,1,10,100},
    ytick={20,40,60,80,100},
    xlabel={False Matching Rate (\%)},
    ylabel={True Matching Rate (\%)},
    tick label style={font=\small},
    axis background/.style={fill=gray!15},
    grid=both,
    grid style={white, line width=0.6pt},
    axis line style={draw=none},
    tick style={draw=black},
    clip=true,
    legend pos=south east,
    legend cell align=left,
    legend style={font=\small, fill=white, draw=gray},
]

\addplot[name path=upper, draw=none, forget plot]
    table[x=far, y expr=\thisrow{tar}+\thisrow{std}] {data/vf-roc-cross-0.dat};
\addplot[name path=lower, draw=none, forget plot]
    table[x=far, y expr=\thisrow{tar}-\thisrow{std}] {data/vf-roc-cross-0.dat};
\addplot[matlightred, opacity=0.25, forget plot]
    fill between[of=upper and lower];
    
\addplot[thick, color=matlightred, mark=none]
    table[x=far, y=tar] {data/vf-roc-cross-0.dat};
\addlegendentry{Newborn}


\addplot[name path=upper, draw=none, forget plot]
    table[x=far, y expr=\thisrow{tar}+\thisrow{std}] {data/vf-roc-cross-1.dat};
\addplot[name path=lower, draw=none, forget plot]
    table[x=far, y expr=\thisrow{tar}-\thisrow{std}] {data/vf-roc-cross-1.dat};
\addplot[matlightblue, opacity=0.25, forget plot]
    fill between[of=upper and lower];
    
\addplot[thick, color=matlightblue, mark=none]
    table[x=far, y=tar] {data/vf-roc-cross-1.dat};
\addlegendentry{7-day}


\addplot[name path=upper, draw=none, forget plot]
    table[x=far, y expr=\thisrow{tar}+\thisrow{std}] {data/vf-roc-cross-2.dat};
\addplot[name path=lower, draw=none, forget plot]
    table[x=far, y expr=\thisrow{tar}-\thisrow{std}] {data/vf-roc-cross-2.dat};
\addplot[matgreen, opacity=0.25, forget plot]
    fill between[of=upper and lower];
    
\addplot[thick, color=matgreen, mark=none]
    table[x=far, y=tar] {data/vf-roc-cross-2.dat};
\addlegendentry{15-day}


\addplot[name path=upper, draw=none, forget plot]
    table[x=far, y expr=\thisrow{tar}+\thisrow{std}] {data/vf-roc-cross-3.dat};
\addplot[name path=lower, draw=none, forget plot]
    table[x=far, y expr=\thisrow{tar}-\thisrow{std}] {data/vf-roc-cross-3.dat};
\addplot[matdarkred, opacity=0.25, forget plot]
    fill between[of=upper and lower];
    
\addplot[thick, color=matdarkred, mark=none]
    table[x=far, y=tar] {data/vf-roc-cross-3.dat};
\addlegendentry{1-month}

\addplot[name path=upper, draw=none, forget plot]
    table[x=far, y expr=\thisrow{tar}+\thisrow{std}] {data/vf-roc-cross-4.dat};
\addplot[name path=lower, draw=none, forget plot]
    table[x=far, y expr=\thisrow{tar}-\thisrow{std}] {data/vf-roc-cross-4.dat};
\addplot[matdarkblue, opacity=0.25, forget plot]
    fill between[of=upper and lower];
    
\addplot[thick, color=matdarkblue, mark=none]
    table[x=far, y=tar] {data/vf-roc-cross-4.dat};
\addlegendentry{2-month}


\addplot[name path=upper, draw=none, forget plot]
    table[x=far, y expr=\thisrow{tar}+\thisrow{std}] {data/vf-roc-cross-5.dat};
\addplot[name path=lower, draw=none, forget plot]
    table[x=far, y expr=\thisrow{tar}-\thisrow{std}] {data/vf-roc-cross-5.dat};
\addplot[orange, opacity=0.25, forget plot]
    fill between[of=upper and lower];
    
\addplot[thick, color=orange, mark=none]
    table[x=far, y=tar] {data/vf-roc-cross-5.dat};
\addlegendentry{3-month}

\end{semilogxaxis}
\end{tikzpicture}
\caption{TMR@FMR curves by age group. Each curve corresponds to the top-performing quality assessment method at FMR=1.0\% for its respective age group.}
\label{fig:res-roc-curves}
\end{figure}
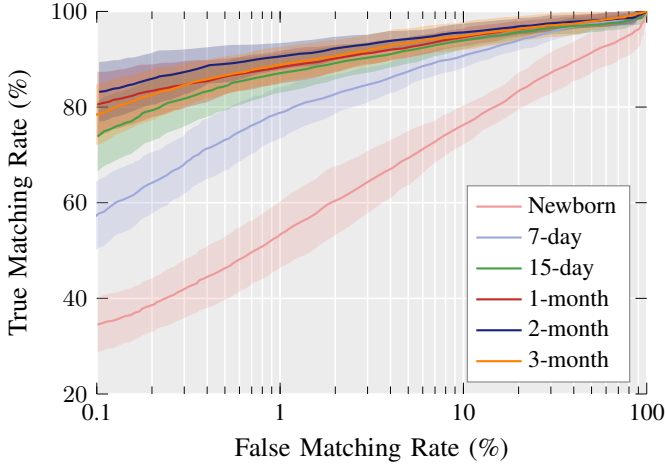

Figure~\ref{fig:res-roc-curves} further illustrates the recognition results with the Receiver Operating Characteristic (ROC) curves for the best-performing methods at FMR=1.0\% (Table~\ref{tab:res-tf-utfpr}), selected separately for each age group. Two key observations follow: (i) newborn and 7-day samples exhibit considerably lower recognition rates than the older age groups, even at the relatively lenient FMR=10\% operating point; and (ii) after 15 days, performance differences become comparatively small, with fluctuations causing the 3-month group to exhibit slightly lower rates despite being comprised of the oldest samples.

\subsection{Improvements to Matching Rates}\label{sec:res-fusion}

In order to improve TMR values, the top-ranked quality methods of Table~\ref{tab:res-tf-utfpr} are combined with a simple frame-fusion strategy. Under this strategy, an ``accept'' requires the probe frame to match at least two frames of the same digits gallery. As such, a genuine probe is compared with the other four frames of its own gallery (all others besides itself), whereas an impostor probe frame is compared with all five gallery frames. This process is less strict than a fusion between two or more digits because it limits combinations to images of the same finger, rather than fingers of the same person.

Additionally, the state-of-the-art comparator MegaMatcher \cite{Neurotechnology2025} (COTS) is employed alongside VeriFinger for latent fingerprint feature extraction and template matching, as comparisons in the literature for similar age groups tend to favor latent solutions \cite{Engelsma2022}, due to their capability of handling degraded impressions. Table~\ref{tab:res-tf-fusion} presents the TMR values achieved by frame-fusion at FMR=1.0\%, using the top ranked quality method of Table~\ref{tab:res-friedman}. Upward arrows denote the relative gain over VeriFinger alone, using the corresponding results reported in Table~\ref{tab:res-tf-utfpr} as reference.

\begin{table}[!htbp]
\centering
\scriptsize
\caption{Matching performance with frame-fusion over VeriFinger and MegaMatcher (Latent) comparators using RPS (tuned) (90 digits per fold, 5 folds).}
\label{tab:res-tf-fusion}
\newcommand{\sd}[1]{{\tiny$\pm$#1}}
\setlength{\tabcolsep}{12pt}
\begin{tabular}{lcc}
\toprule
& \textbf{VeriFinger} & \textbf{MegaMatcher} \\
\textbf{Age} & \textbf{TMR@FMR=1.0\%} & \textbf{TMR@FMR=1.0\%} \\
\midrule
Newborn & 60.1 \sd{7.8} {\scriptsize($\uparrow 12.6$)} & 77.2 \sd{6.4} {\scriptsize($\uparrow 29.7$)} \\
\midrule
7-day & 85.6 \sd{2.6} {\scriptsize($\uparrow 08.3$)} & 93.2 \sd{1.4} {\scriptsize($\uparrow 15.9$)} \\
\midrule
15-day & 92.1 \sd{3.1} {\scriptsize($\uparrow 04.8$)} & 96.4 \sd{1.6} {\scriptsize($\uparrow 09.1$)} \\
\midrule
1-month & 92.8 \sd{2.5} {\scriptsize($\uparrow 04.4$)} & 96.4 \sd{2.8} {\scriptsize($\uparrow 08.0$)} \\
\midrule
2-month & 94.5 \sd{1.0} {\scriptsize($\uparrow 04.0$)} & 97.6 \sd{0.9} {\scriptsize($\uparrow 07.1$)} \\
\midrule
3-month & 93.9 \sd{2.8} {\scriptsize($\uparrow 05.1$)} & 97.5 \sd{1.6} {\scriptsize($\uparrow 08.7$)} \\
\bottomrule
\end{tabular}
\end{table}

As shown, frame-fusion improved upon the corresponding single-frame results of Table~\ref{tab:res-tf-utfpr} in every age group, and latent matching also obtained higher TMR than the Verifinger baseline for all cases. For the 2-month group, fusion and latent matching elevated the top performance from a $90.5\%$ mean (single-frame, VeriFinger) to $97.6\%$, the highest mean value encountered across tables, with a standard deviation of only $0.9$ points. Newborns obtained the largest gains ($29.7$ points with MegaMatcher; e.g., RPS from $47.5\%$ to $77.2\%$) from this strategy, but remained the most challenging group as MegaMatcher TMR mean remains at least $16$ points below the other age groups.

From 15 days onward, MegaMatcher TMR exceeded $96\%$ for all listed methods, which supports video sampling as a viable strategy for capturing high-utility fingerprints from the earliest stages of life, and is compatible with the tuned methods selecting useful frames. Since Table~\ref{tab:res-tf-fusion} does not include default or NFIQ2-selected galleries, this experiment does not isolate the contribution of tuning to the fusion gains.

\begin{figure}[!htb]
\centering
\input{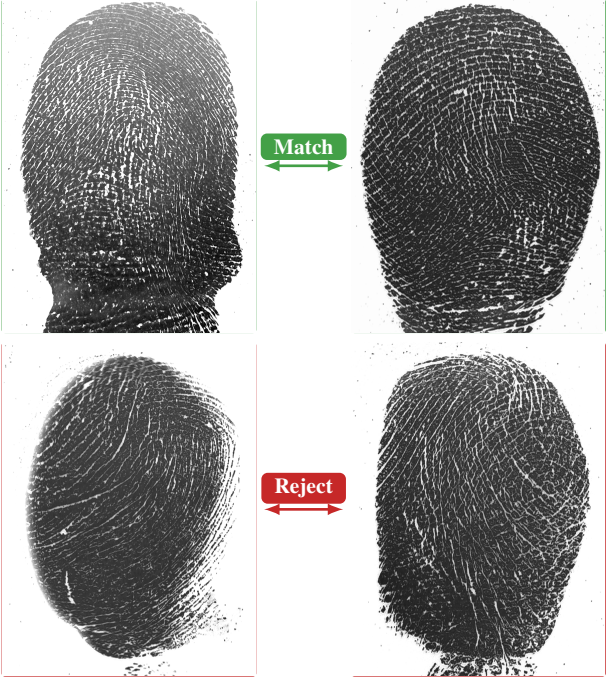}
\caption{Examples of genuine newborn sample pairs. The thumb pair (top) exceeded the MegaMatcher threshold, whereas the little-finger pair (bottom) did not.}
\label{fig:res-pairs}
\end{figure}

Nevertheless, challenges remain in reliably capturing and evaluating fingerprints during the first week after birth, including verification at maternity. Even with thousands of video frames available, 100\% TMR may not be achievable at reasonable FMR operating points without a Failure-To-Enroll (FTE) rule. Real-time quality evaluation and feedback to collection agents could therefore help identify cases requiring additional acquisitions. Figure~\ref{fig:res-pairs} illustrates this limitation using two genuine newborn sample pairs: a successfully matched thumb pair (top) and a false reject involving the little fingers (bottom). The pairs were selected using RPS quality scores and evaluated with frame-fusion and latent matching with a threshold of 70, which yields FMR=1.0\% for RPS.

Although the thumb samples are visually more distinct, involuntary movement, acquisition noise, and the lack of suitable pairings can prevent successful matching, even for relatively clear frames such as the bottom-left image. In such cases, low-quality samples could be excluded using quality-score thresholds that balance TMR gains against exclusion rates, as observed during the FNMR validation in Figure~\ref{fig:bayes-fnmr}.

\section{Discussion}\label{sec:dis}

The comparison between default and tuned configurations indicates that parameters derived from adult fingerprint norms limit the applicability of otherwise suitable methods to infants. At FMR=1.0\%, tuned configurations obtained higher mean TMR than their defaults in every method-age combination. The effect is clearest for the methods with strong fingerprint-specific assumptions, LCS and RPS, whose rank gains over their defaults exceeded the CD at both operating points. Together, these two methods obtained the highest mean TMR at FMR=1.0\% in five of six age groups, with margins within the fold standard deviation, however neither  differed significantly from the other top-five methods. 

Nevertheless, not all methods benefited equally from tuning. Gabor results were mixed, with consistent gains at FMR=1.0\% but not FMR=0.1\%. The simpler block-size tuning produced varying gains for RVU and OCL, but neither method was lifted above the adult-oriented standard NFIQ2. Methods which have no tunable inputs, namely SIG and MQCDA, were also shown to be competitive, ranking among the top five at FMR=1.0\%, and together obtained the highest TMR mean at FMR=0.1\% for three out of the six observed age groups.

In regards to subject age, matching rates was notably lower for newborns, consistent with the well-established difficulty of acquiring high-quality fingerprint images at this age. Performance was also shown to generally improve with age, with only minor variations between the 15-day and 1-month groups and between the 2- and 3-month groups. Within the evaluated subset of UTFPR-IFD, the 2-month group reached the highest TMR mean in frame comparisons, $90.5\%$ (RPS), which rose to $97.6$ once frame-fusion was combined with a COTS latent comparator (Table~\ref{tab:res-tf-fusion}). Critically, differences across the 15-day to 3-month age groups ($97.4$--$97.5\%$) are within the fold standard deviation. Further improvements could be obtained as this fusion rule, which requires two matching frames of the same finger, is less strict than multi-finger acceptance protocols, which could yield further gains.

Finally, the Friedman–Nemenyi analysis does not identify a single superior method among the top five, as the differences in their average ranks do not exceed the critical difference. However, these methods rank consistently above most of the remaining methods and configurations, which suggests that they are the most appropriate solutions for infant fingerprints. A combinatorial approach akin to NFIQ2, assimilating complementary ridge properties such as orientation, width and frequency into a unified quality index, therefore appears to be the most promising direction for practical infant fingerprint quality assessment.

In summary, the findings yield several conclusions:
\begin{itemize}
    \item Infant fingerprint patterns differ substantially from adults, such that standard parameters (devised for adults) are ill-suited to capture the characteristic ridge widths, frequencies, and structural properties of infant samples;
    \item Parameterized quality assessment methods can be adapted to infant fingerprints through parameter search, and the resulting configurations establish a baseline for future comparisons;
    \item Matching rates remain relatively low for infants in sample-to-sample comparisons. However, frame fusion and latent comparators substantially improve results. Newborns nevertheless remain the most challenging age group as Table~\ref{tab:res-tf-fusion} reports;
    \item A composite index, such as NFIQ2, that integrates complementary features and extends this concept to incorporate age-adaptive components, represents a promising direction for quality-guided infant biometric enrollment.
\end{itemize}

\section{Conclusions}\label{sec:con}

This study examined the applicability of available fingerprint quality assessment methods for infants, initially enrolled as newborns and accompanied until 3-month of age. The majority of these methods were designed around adult fingerprint characteristics and are currently part of an international standard \cite{ISO}. A parameter search strategy was proposed to reduce the gap between adult and infant fingerprint recognition, using a longitudinal dataset that tracks subjects from birth onward~\cite{Southier2025:base}. The final parameter set produced in this work (separate execution, full database) is available as supplementary material in Appendix~\ref{app:parameter-set}.

By directing the parameter search toward recognition performance, optimal ranges for algorithm features, including ridge and valley width, were identified. These tuned parameters produced measurable improvements in matching rates (TMR@FMR) across all evaluated age groups, demonstrating that parameters derived from adult standards are insufficient when applied to infant fingerprints. Additionally, a stricter recognition protocol could yield further improvements in matching rates, such as requiring the successful match of two or more fingers as recommended by ISO/IEC 2382-37~\cite{ISO2022}. Section~\ref{sec:res-fusion} explores a similar frame-fusion solution that is shown to substantially improve TMR values across all ages.

Collectively, these results indicate that existing international standards do not adequately account for the biometric characteristics of infant fingerprints, which differ measurably from those of adults in feature scale and distribution. Nevertheless, the proposed parameter search demonstrates that existing biometric frameworks can be adapted to infant subjects when representative data is available. The optimized parameter sets identified in this work provide a practical foundation for future work aiming to propose solutions to infant fingerprint recognition in maternity settings, where reliable biometric validation helps ensure the identity of newborns to their parents.

Several directions remain open for future research: further investigation into how aging affects ridge and valley widths beyond the observed age groups, potentially up to adulthood; the proposal of international standards that explicitly account for age-dependent pattern characteristics and resolution variability; and the proposal of composite quality indexes for infants. The development of quality estimation, enhancement, feature extraction, and matching methods for high-resolution fingerprints could also yield significant improvements for infant biometrics.

\section{Acknowledgements}

We sincerely thank InfantID, CAPES (Finance Code 001), CNPq, Araucária Foundation, and FINEP, for their financial support.

\footnotesize
\bibliographystyle{IEEEtran}
\bibliography{bibliography}

\clearpage
\appendices

\section{Complete parameter set}\label{app:parameter-set}

\begin{table}[!htb]
\caption{Tuned parameter sets (Part I).}
\label{tab:combined-parameters-pt1}
\centering
\scriptsize
\rowcolors{1}{white}{gray!15}
\begin{tabular}{l l l l}
\hline
Age Group & 
\textbf{Gabor} & 
\textbf{LCS} &
\textbf{RVU} \\
\cline{2-4}
&
\textbf{($m$, $f$, $\sigma$, $\sigma_g$)} &
\textbf{$Wr$---$Wv$, B. Size} &
\textbf{B. Size} \\
\hline
Newborn &
8, 0.500, 0.807, 1 &
2--11, 1--3, $64\times64$ &
$128\times128$ \\

7-day &
8, 0.500, 0.807, 5 &
3--6, 1--3, $64\times64$ &
$64\times64$ \\

15-day &
8, 0.050, 8.071, 5 &
2--14, 2--3, $64\times64$ &
$64\times64$ \\

1-month &
8, 0.500, 0.807, 1 &
2--9, 2--5, $64\times64$ &
$64\times64$ \\

2-month &
8, 0.500, 0.807, 2 &
2--9, 2--5, $64\times64$ &
$64\times64$ \\

3-month &
8, 0.500, 0.807, 1 &
2--9, 1--6, $64\times64$ &
$64\times64$ \\
\hline
Adult &
4, 0.100, 6.000, 1~\cite{ISO} &
3--10, 2--10, $32\times32$~\cite{ISO} &
$64\times64$~\cite{ISO} \\
\hline
\end{tabular}
\end{table}

\begin{table}[!htb]
\caption{Tuned parameter sets (Part II).}
\label{tab:combined-parameters-pt2}
\centering
\scriptsize
\rowcolors{1}{white}{gray!15}
\begin{tabular}{l l l l}
\hline
Age Group & 
\textbf{RPS} &
\textbf{OCL} &
\textbf{DFB} \\
\cline{2-4}
&
\textbf{($r_{min}$--$r_{max}$, $\Delta r$, $\theta$)} &
\textbf{B. Size} &
\textbf{B. Size} \\
\hline
Newborn &
0.211--0.397, 10, 100 &
$128\times128$ &
$64\times64$ \\

7-day &
0.379--0.560, 22, 79 &
$128\times128$ &
$64\times64$ \\

15-day &
0.440--0.700, 17, 132 &
$128\times128$ &
$64\times64$ \\

1-month &
0.240--0.700, 57, 132 &
$128\times128$ &
$64\times64$ \\

2-month &
0.440--0.700, 81, 81 &
$128\times128$ &
$64\times64$ \\

3-month &
0.440--0.700, 2, 114 &
$128\times128$ &
$64\times64$ \\
\hline
Adult &
0.077--0.143, 10, 180~\cite{Olsen2016, ISO} &
$32\times32$~\cite{ISO} &
$4\times4$~\cite{Hendre2022} \\
\hline
\end{tabular}
\end{table}

\end{document}